%% file: paper.tex
\documentclass{article} % For LaTeX2e
\usepackage{hyperref}
\usepackage{url}
\usepackage[english]{babel}
\usepackage[utf8]{inputenc}  % in this day and age, we need utf8 and 8-bit fonts.
\usepackage[T1]{fontenc}

\usepackage{iclr2019_conference,times}

\usepackage{algorithm}
\usepackage{algorithmic}

\usepackage{import}
\usepackage{graphicx}
\usepackage{amssymb,amsmath,amsfonts}
\usepackage{enumitem}   
\usepackage{bm}
\usepackage{cancel}

\DeclareMathOperator*{\E}{\mathbb{E}}
\DeclareMathOperator*{\Var}{\mathbb{V}}
\DeclareMathOperator*{\Cov}{\mathbb{C}}

\newcommand{\Normal}[1]{\mathcal{N}\left(#1\right)}

\newcommand{\ssup}[1]{^{(#1)}}
\newcommand{\xind}{x}
\newcommand{\yind}{y}
\newcommand{\bracket}[3]{\left#1 #3 \right#2}
\newcommand{\bra}{\bracket{(}{)}}

\newcommand{\sqb}{\bracket{[}{]}}

\newcommand{\vx}{{\bf {x}}}

\newcommand{\vX}{{\bf {X}}}
\newcommand{\vY}{{\bf {Y}}}
\newcommand{\vW}{{\bf {W}}}

\newcommand{\vU}{{\bf {U}}}

\newcommand{\0}{{\bf {0}}}

\newcommand{\vA}{{\bf {A}}}

\newcommand{\vK}{{\bf {K}}}

\newcommand{\vb}{{\bf {b}}}

\newcommand{\va}{{\bf {a}}}

\newcommand{\eqdef}{:=}

\usepackage[single=true]{acro}
\DeclareAcronym{NN}{
  short = NN,
  long  = neural network,
  class = abbrev
}
\DeclareAcronym{CNN}{
  short = CNN,
  long  = convolutional neural network,
  class = abbrev
}
\DeclareAcronym{CLT}{
  short = CLT,
  long  = Central Limit Theorem,
  class = abbrev
}
\DeclareAcronym{RV}{
  short = RV,
  long  = random variable,
  long-plural-form = random variables,
  class = abbrev
}
\DeclareAcronym{iid}{
  short = iid,
  long  = independent and identically distributed,
  class = abbrev
}
\DeclareAcronym{GP}{
  short = GP,
  long  = Gaussian process,
  long-plural-form = Gaussian Processes,
  class = abbrev
}

\usepackage{lscape}

\title{Deep Convolutional Networks as\\shallow Gaussian Processes}

\author{
  Adrià Garriga-Alonso \\
  department of Engineering \\
  University of Cambridge \\
  \texttt{ag919@cam.ac.uk}
  \And
  Carl Edward Rasmussen\\
  Department of Engineering \\
  University of Cambridge \\
  \texttt{cer54@cam.ac.uk}
  \And
  Laurence Aitchison\\
  Department of Engineering \\
  University of Cambridge \\
  \texttt{laurence.aitchison@gmail.com}
}

\iclrfinalcopy % Uncomment for camera-ready version, but NOT for submission.
\begin{document}

\maketitle

\begin{abstract}
We show that the output of a (residual) \ac{CNN} with an appropriate prior over the weights and biases is a \ac{GP} in the limit of infinitely many convolutional filters,
extending similar results for dense networks. 
For a \ac{CNN}, the equivalent kernel can be computed exactly and, unlike ``deep kernels'', has very few parameters: only the hyperparameters of the original \ac{CNN}.
Further, we show that this kernel has two properties that allow it to be computed efficiently; the cost of evaluating the kernel for a pair of images is similar to a single forward pass through the original \ac{CNN} with only one filter per layer. 
The kernel equivalent to a 32-layer ResNet obtains 0.84\% classification error on MNIST, a new record for \acp{GP} with a comparable number of parameters.
\footnote{Code to replicate this paper is available at \href{https://github.com/convnets-as-gps/convnets-as-gps}{https://github.com/convnets-as-gps/convnets-as-gps}}
\end{abstract} 

\section{Introduction}

Convolutional Neural Networks (\acp{CNN}) have powerful pattern-recognition
capabilities that have recently given dramatic improvements in important tasks
such as image classification \citep{krizhevsky2012imagenet}.
However, as \acp{CNN} are increasingly being applied in real-world, safety-critical domains, their vulnerability to adversarial examples \citep{szegedy2013intriguing,kurakin2016adversarial}, and their poor uncertainty estimates are becoming increasingly problematic.
Bayesian inference is a theoretically principled and demonstrably successful \citep{snoek2012practical,deisenroth2011pilco} framework for learning in the face of uncertainty, which may also help to address the problems of adversarial examples \citep{gal2018idealised}.
Unfortunately, Bayesian inference in \acp{CNN} is extremely difficult due to the very large number of parameters, requiring highly approximate factorised variational approximations \citep{blundell2015weight,gal2015dropout}, or requiring the storage \citep{lakshminarayanan2017simple} of large numbers of posterior samples \citep{welling2011bayesian,mandt2017stochastic}.

Other methods such as those based on \acfp{GP} are more amenable to Bayesian inference,
allowing us to compute the posterior uncertainty exactly
\citep{rasmussen2006gaussian}. This raises the question of whether it might be
possible to combine the pattern-recognition capabilities of \acp{CNN} with the exact probabilistic computations in \acp{GP}.
Two such approaches exist in the literature.
First, deep convolutional kernels \citep{wilson2016deep} parameterise a \ac{GP} kernel using the weights and biases of a \ac{CNN}, which is used to embed the input images into some latent space before computing their similarity. 
The \ac{CNN} parameters of the resulting kernel then are optimised by gradient descent. 
However, the large number of kernel parameters in the \ac{CNN} reintroduces the
risk of overconfidence and overfitting. To avoid this risk, we need to infer a posterior over the \ac{CNN} kernel parameters, which is as difficult as directly inferring a posterior over the parameters of the original \ac{CNN}.
Second, it is possible to define a convolutional \ac{GP} \citep{markvdw2017convolutional} or a deep convolutional \ac{GP} \citep{kumar2018deep} by defining a \ac{GP} that takes an image patch as input, and using that \ac{GP} as a component in a larger \ac{CNN}-like system.
However, inference in such systems is very computationally expensive, at least without the use of potentially severe variational approximations \citep{markvdw2017convolutional}.

An alternative approach is suggested by the underlying connection between
Bayesian Neural Networks (\acsp{NN}) and \acp{GP}.
In particular, \citet{neal1996bayesian} showed that the function defined by a single-layer fully-connected \ac{NN} with infinitely many hidden units, and random independent zero-mean weights and biases is equivalent to a \ac{GP}, implying that we can do exact Bayesian inference in such a \ac{NN} by working with the equivalent \ac{GP}. 
%\footnote{Should I hedge this by saying it ``approximates a GP as the number of hidden units goes to infinity''?}
%\citep{neal1996bayesian}. 
%The posterior and predictive distributions for a \ac{GP} prior can be computed exactly for regression, albeit in $O(N^3)$ time for $N$ training examples \citep{rasmussen2006gaussian}. 
%There also exist a wealth of approximation schemes that can be used for classification or for reducing the $O(N^3)$ time cost \citep{Bui2017AUF,hensman-svgp}.
Recently, this result was extended to arbitrarily deep fully-connected \acp{NN} with infinitely many hidden units in each layer \citep{lee2017deep,
gpbehaviour}.
However, these fully-connected networks are rarely used in practice, as they are unable to exploit important properties of images such as translational invariance, raising the question of whether state-of-the-art architectures such as \acp{CNN} \citep{lecun1990handwritten} and ResNets \citep{he2016deep} have equivalent \ac{GP} representations.
Here, we answer in the affirmative, giving the \ac{GP} kernel corresponding to arbitrarily deep \acp{CNN} and to (convolutional) residual neural networks \citep{he2016deep}.
In this case, if each hidden layer has an infinite number of convolutional \emph{filters}, the network prior is equivalent to a \ac{GP}. 

Furthermore, we show that two properties of the \ac{GP} kernel induced by a \ac{CNN} allow it to be computed very efficiently.
First, in previous work it was necessary to compute the covariance matrix for the output of a single convolutional filter applied at all possible locations within a single image \citep{markvdw2017convolutional}, which was prohibitively computationally expensive.
In contrast, under our prior, the downstream weights are independent with zero-mean, which decorrelates the contribution from each location, and implies that it is necessary only to track the patch variances, and not their covariances.
Second, while it is still necessary to compute the variance of the output of a convolutional filter applied at all locations within the image, the specific structure of the kernel induced by the \ac{CNN} means that the variance at every location can be computed simultaneously and efficiently as a convolution.

Finally, we empirically demonstrate the performance increase coming from adding translation-invariant structure to the \ac{GP} prior. 
Without computing any gradients, and without augmenting the training set (e.g.\
using translations), we obtain 0.84\% error rate on the MNIST classification
benchmark, setting a new record for nonparametric GP-based methods.

\section{\ac{GP} behaviour in a \ac{CNN}}

For clarity of exposition, we will treat the case of a 2D convolutional \ac{NN}.
The result applies straightforwardly to $n$D convolutions, dilated convolutions
and upconvolutions (``deconvolutions''), since they can be represented as linear
transformations with tied coefficients (see Fig.~\ref{fig:conv-as-linear}).

\subsection{A 2D convolutional network prior\label{sec:define-convnet}}
The network takes an arbitrary input image $\vX$ of height $H\ssup{0}$ and
width $D\ssup{0}$, as a $C\ssup{0} \times (H\ssup{0}
D\ssup{0})$ real matrix. 
% Let $\vX \in \reals^{C^{(0)} \times (H^{(0)} \cdot W^{(0)})}$ be an input
% image to the network.
Each row, which we denote $\vx_1,\vx_2,\dotsc,\vx_{C\ssup{0}}$, corresponds to a channel
of the image (e.g. $C\ssup{0} = 3$ for RGB), flattened to form a 
vector.
% \footnote{Is it weird that in this notation rows of the matrix become column vectors?}
% Each channel is ``flattened'' to form a column vector $\vx_i$.
The first activations $\vA\ssup{1}(\vX)$
are a linear transformation of the inputs. For $i \in \{1,\dotsc,C\ssup{1}\}$:
\begin{equation}
  \va_i\ssup{1}(\vX) \eqdef b_i\ssup{1} {\bf 1} + \sum_{j=1}^{C\ssup{0}} \vW\ssup{1}_{i,j} \vx_j\;.
\label{eq:network-base}
\end{equation}

We consider a network with $L$ hidden layers. The other activations of the
network, from $\vA\ssup{2}(\vX)$ up to $\vA\ssup{L+1}(\vX)$, are defined
recursively:
\begin{equation}
  \va_i\ssup{\ell+1}(\vX) :=  b_i\ssup{\ell+1} {\bf 1} + \sum_{j=1}^{C\ssup{\ell}}  \vW_{i,j}\ssup{\ell+1} \phi\left( \va_j\ssup{\ell}(\vX) \right).
\label{eq:network-recursive}
\end{equation}

The activations $\vA\ssup{\ell}(\vX)$ are $C\ssup{\ell} \times (H\ssup{\ell}
D\ssup{\ell})$ matrices. Each row $\va_i\ssup{\ell+1}$ represents
the flattened $j$th channel of the image that results
from applying a convolutional filter to $\phi(\vA\ssup{\ell}(\vX))$.

\begin{figure}%[ht]
  \centerline{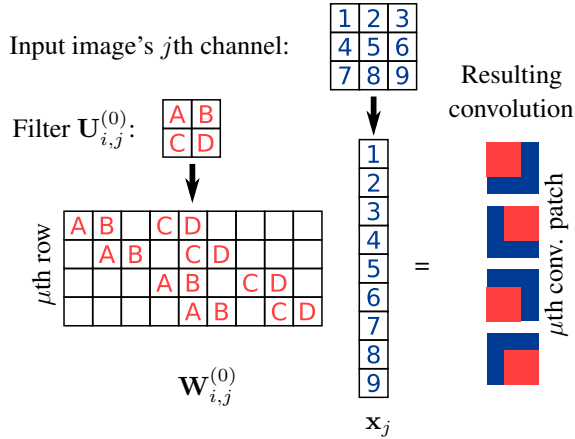}
  \caption{The 2D convolution $\vU\ssup{0}_{i,j} * \vx_j$ as the dot product
  $\vW\ssup{0}_{i,j} \vx_j$. The blank elements of $\vW\ssup{0}_{i,j}$ are zeros.
  The $\mu$th row of $\vW\ssup{0}_{i,j}$ corresponds to applying the filter to the
  $\mu$th convolutional patch of the channel $\vx_j$. \label{fig:conv-as-linear}}
\end{figure}

The structure of the pseudo-weight matrices
%$\Big\{\Big\{\vW^{(\ell)}_{j,i}\Big\}_{i=1}^{C^{(\ell)}}\Big\}_{j=1}^{C^{(\ell+1)}}$
$\vW\ssup{\ell+1}_{i,j}$
and biases $b_i\ssup{\ell+1}$, for $i\in \{1,\dotsc,C\ssup{\ell+1}\}$ and $j \in \{1,\dots,C\ssup{\ell}\}$,
%$\Big\{\vb^{(\ell)}_j\Big\}_{j=1}^{C^{(\ell + 1)}}$
%$\{\vb^{(\ell)}_j\}_{j=1}^{C^{(\ell + 1)}}$
% where $i$ ranges from $1$ to $C\ssup{\ell}$ and $j$ ranges from $1$ to $C\ssup{\ell + 1}$
depends on the architecture. For a convolutional layer, each row of
$\vW\ssup{\ell+1}_{i,j}$ represents a \emph{position} of the filter, such that the dot
product of all the rows with the image vector $\vx_j$ represents applying the
convolutional filter $\vU\ssup{\ell+1}_{i,j}$ to the $j$th channel. Thus, the elements of each row of
$\vW\ssup{\ell+1}_{i,j}$ are: 0 where the filter does not apply, and the corresponding
element of $\vU_{i,j}\ssup{\ell+1}$ where it does, as illustrated in
Fig.~\ref{fig:conv-as-linear}. 
%For the $g$th position of the filter, 
%the elements of the previous layer's
%activation where the filter applies form the $g$th
%\emph{convolutional patch}.

%If the bias is shared among filter positions, all the
%elements of $\vb_j^{(\ell)}$ have the same value, $b_j^{(\ell)}$.

The outputs of the network are the last activations, $\vA\ssup{L+1}(\vX)$. 
In the classification or regression setting, the outputs are not spatially extended, so we have $H\ssup{L+1} = D\ssup{L+1} = 1$, which is equivalent to a fully-connected output layer.
%We can make layer $L+1$ fully
%connected (typically, only the last layer) just by considering convolutional
%filters with as many elements as the $\ell$th activation.
In this case, the pseudo-weights $\vW\ssup{L+1}_{i,j}$ only have one
row, and the activations $\va_i\ssup{L+1}$ are single-element vectors.
%If we're performing classification or regression, there is 
%one final
%activation $a_j\ssup{L+1}$ for each of the regression outputs, or for each class to classify.

Finally, we define the prior distribution over functions by making the filters
$\vU_{i,j}\ssup{\ell}$ and biases $b_i\ssup{\ell}$ be independent Gaussian \acp{RV}. For
each layer $\ell$, channels $i,j$ and locations within the filter $\xind,\yind$:
\begin{align}
  \label{eq:prior}
  U\ssup{\ell}_{i,j,\xind,\yind} &\sim \Normal{0, \sigma_\text{w}^2/C\ssup{\ell}}, & b\ssup{\ell}_i &\sim \Normal{0, \sigma_\text{b}^2}.
\end{align}
Note that, to keep the activation variance constant, the weight variance is
divided by the number of input channels. The weight variance can also be divided
by the number of elements of the filter, which makes it equivalent to the
\ac{NN} weight initialisation scheme introduced by \citet{he2016deep}.

% Optionally, we could also divide $\sigma_w^2$ at layer $\ell$ by the size of the
% filter at layer $\ell$, to prevent the variance from growing with it.

% \subsection{Independent random filters define a \ac{GP}}

% The same happens when using a \ac{GP} model with any kernel: the prior outputs
% for regression follow a \ac{GP} but the prior for classification can't, since
% it's a distribution over 
% This is exactly analogous to normal \ac{GP} usage: regression is conjugate but
% classification isn't.

% \begin{figure}[h]
% \vspace{.3in}
% \centerline{\includegraphics[width=0.5\textwidth]{img/convnet.png}}
% \vspace{.3in}
% \caption{Diagram of the \ac{CNN} of interest, with components labelled according
%   to our notation.}
% \end{figure}

\subsection{Argument for \ac{GP} behaviour\label{sec:argument-gp}}

We follow the proofs by \citet{lee2017deep} and \citet{gpbehaviour} to show that the output of the
\ac{CNN} described in the previous section, $\vA\ssup{L+1}$, defines a \ac{GP}
indexed by the inputs, $\vX$.
Their proof \citep{lee2017deep} proceeds by applying the multivariate \ac{CLT} to each layer in sequence, i.e. taking the limit as $N\ssup{1} \rightarrow \infty$, then $N\ssup{2} \rightarrow \infty$ etc, where $N\ssup{\ell}$ is the number of hidden units in layer $\ell$.
By analogy, we sequentially apply the multivariate CLT by taking the limit as the number of channels goes to infinity, i.e. $C\ssup{1} \rightarrow \infty$, then $C\ssup{2} \rightarrow \infty$ etc.
While this is the simplest approach to taking the limits, other potentially more
realistic approaches also exist \citep{gpbehaviour}. In
Appendix~\ref{sec:actual-proof} we take the latter approach.

The fundamental quantity we consider is a vector formed by concatenating the feature maps (or equivalently channels), $\va_j\ssup{\ell}(\vX)$ and $\va_j\ssup{\ell}(\vX')$ from data points $\vX$ and $\vX'$,
\begin{equation}
  \label{eq:vapair}
  \va_i\ssup{\ell}(\vX, \vX') = 
  \begin{pmatrix} \va_i\ssup{\ell}(\vX)\phantom{'} \\ \va_i\ssup{\ell}(\vX') \end{pmatrix}.
\end{equation}
This quantity (and the following arguments) can all be extended to the case of
finitely many input points.

\paragraph{Induction base case.}
For any pair of data points, $\vX$ and $\vX'$ the feature-maps corresponding to
the $j$th channel, $\va_j\ssup{1}(\vX, \vX')$ have a multivariate Gaussian joint
distribution. This is because each element is a linear combination of shared
Gaussian random variables: the biases, $\vb_j\ssup{0}$ and the filters,
$\vU\ssup{0}_{j,:}$. Following Eq.~\eqref{eq:network-base},
\begin{align}
  \va_i^{(1)}(\vX, \vX')
  &= b_i^{(1)} {\bf 1}
     + \sum_{i=1}^{C\ssup{0}} 
     \begin{pmatrix} \vW_{i,j}\ssup{1} & \0  \\ \0& \vW_{i,j}\ssup{1} \end{pmatrix}
     \begin{pmatrix} \vx_i \\ \vx_i' \end{pmatrix},
\end{align}
where ${\bf 1}$ is a vector of all-ones.
While the elements within a feature map display strong correlations, different feature maps are \ac{iid} conditioned on the data (i.e. $\va_i\ssup{1}(\vX,\vX')$ and $\va_{i'}\ssup{1}(\vX,\vX')$ are \ac{iid} for $i\neq i'$), because the parameters for different feature-maps (i.e. the biases, $b_i\ssup{1}$ and the filters, $\vW\ssup{1}_{i,:}$) are themselves \ac{iid}.

\paragraph{Induction step.}
Consider the feature maps at the $\ell$th layer, $\va_j\ssup{\ell}(\vX, \vX')$, to be \ac{iid} multivariate Gaussian \acp{RV} (i.e. for $j\neq j'$, $\va_{j}\ssup{\ell}(\vX, \vX')$ and $\va_{j'}\ssup{\ell}(\vX, \vX')$ are \ac{iid}).
Our goal is to show that, taking the number of channels at layer $\ell$ to infinity (i.e. $C\ssup{\ell} \rightarrow \infty$), the same properties hold at the next layer (i.e. all feature maps, $\va_i\ssup{\ell+1}(\vX,\vX')$, are \ac{iid} multivariate Gaussian \acp{RV}).
Writing Eq.~\eqref{eq:network-recursive} for two training examples, $\vX$ and $\vX'$, we obtain,
\begin{align}
  \label{eq:network-recursive-pair}
  \va_i\ssup{\ell+1}(\vX,\vX')
  &= b_i\ssup{\ell+1} {\bf 1} 
     + \sum_{j=1}^{C\ssup{\ell}} \begin{pmatrix} \vW_{i,j}\ssup{\ell+1} & \0 \\ \0 & \vW_{i,j}\ssup{\ell+1} \end{pmatrix} 
  \phi(\va_j\ssup{\ell}(\vX,\vX'))
\end{align}
We begin by showing that $\va_i\ssup{\ell+1}(\vX,\vX')$ is a multivariate Gaussian \ac{RV}.
The first term is multivariate Gaussian, as it is a linear function of $b_i\ssup{\ell+1}$, which is itself \ac{iid} Gaussian.
We can apply the multivariate \ac{CLT} to show that the second term is also Gaussian, because, in the limit as $C\ssup{\ell} \rightarrow \infty$, it is the sum of infinitely many \ac{iid} terms: $\va_j^{(\ell)}(\vX,\vX')$ are \ac{iid} by assumption, and $\vW_{i,j}\ssup{\ell+1}$ are \ac{iid} by definition.
%For the second term, note that all the terms in the sum, indexed $i$, are independent, as the 
%Therefore, as we take the number of channels to infinity, and scale appropriately \citep{gpbehaviour}, the multivariate \ac{CLT} applies, so $\va_j^{(\ell+1)}(\vX,\vX')$ is indeed multivariate Gaussian.
Note that the same argument applies to all feature maps jointly, so all elements of $\vA\ssup{\ell+1}(\vX,\vX')$ (defined by analogy with Eq.~\ref{eq:vapair}) are jointly multivariate Gaussian.

Following \citet{lee2017deep}, to complete the argument, we need to show that the output feature maps are \ac{iid}, i.e. $\va_i\ssup{\ell+1}(\vX,\vX')$ and $\va_{i'}\ssup{\ell+1}(\vX,\vX')$ are \ac{iid} for $i\neq i'$.
They are identically distributed, as $b_i\ssup{\ell+1}$ and $\vW_{i, j}\ssup{\ell+1}$ are \ac{iid} and $\phi(\va_j\ssup{\ell}(\vX,\vX'))$ is shared. 
To show that they are independent, remember that $\va_i\ssup{\ell+1}(\vX,\vX')$
and $\va_{i'}\ssup{\ell+1}(\vX,\vX')$ are jointly Gaussian, so it is sufficient
to show that they are uncorrelated. We can show that they are uncorrelated
noting that the weights $\vW_{i, j}\ssup{\ell+1}$ are independent with
zero-mean, eliminating any correlations that might arise through the shared
random vector, $\phi(\va_j\ssup{\ell}(\vX,\vX'))$.

\section{The ConvNet and ResNet kernels}
Here we derive a computationally efficient kernel corresponding to the \ac{CNN} described in the previous section.
It is surprising that we can compute the kernel efficiently because the feature maps, $\va_i\ssup{\ell}(\vX)$, display rich covariance structure due to the shared convolutional filter. Computing and representing these covariances would be prohibitively computationally expensive.
However, in many cases we only need the variance of the output,
e.g. in the case of classification or regression with a final dense layer.
It turns out that this propagates backwards through the convolutional network,
implying that for every layer, we only need the ``diagonal covariance'' of the
activations: the covariance between the corresponding elements of
$\va_j\ssup{\ell}(\vX)$ and $\va_j\ssup{\ell}(\vX')$, that is, $\operatorname{diag}\bra{\Cov\sqb{\va_j\ssup{\ell}(\vX), \va_j\ssup{\ell}(\vX')}}$.

%In the case of
%multi-output regression or classification, there are no other
%covariances: the activations $a_j^{(L+1)}(\vX)$ are \ac{iid} for each $j$, and
%also scalars.
%[I know what you mean, but it's unnecessary for the moment.]

%However, if the outputs are vectors (e.g. if the last few layers are
%upconvolutions), we probably want the covariance between elements of the same
%activation $\va_j^{(L+1)}(\vX)$. We have shown that the outputs in this case
%also follow a \ac{GP}, but we do not give the expression for the
%within-activation covariances in this article.
%[I'm not actually sure what this means...]

\subsection{\ac{GP} mean and covariance\label{sec:gp-covariance}}
A \ac{GP} is completely specified by its mean and covariance (or kernel) functions.
These give the parameters of the joint Gaussian distribution of the \acp{RV} indexed by any two
inputs, $\vX$ and $\vX'$.
For the purposes of computing the mean and covariance, it is easiest to consider the network as being written entirely in index notation,
\begin{equation*}
  A_{i,\mu}\ssup{\ell+1}(\vX) = b_i\ssup{\ell+1} +
    \sum_{j=1}^{C\ssup{\ell}} \ \sum_{\nu=1}^{H\ssup{\ell}\!D\ssup{\ell}} W\ssup{\ell+1}_{i,j,\mu,\nu}\, \phi(A\ssup{\ell}_{j,\nu}(\vX)).
\end{equation*}
where $\ell$ and $\ell+1$ denote the input and output layers respectively, $j$ and $i \in \{1,\dotsc,C\ssup{\ell+1}\}$ denote the input and output channels, and $\nu$ and $\mu \in \{1,\dotsc,H\ssup{\ell+1} D\ssup{\ell+1}\}$ denote the location within the input and output channel or feature-maps.

The mean function is thus easy to compute,
\begin{equation*}
  \E\sqb{A_{i,\mu}\ssup{\ell+1}(\vX)} = \E\sqb{b_i\ssup{\ell+1}} +
    %\sum_{i=1}^{C\ssup{L}}  \E\left[\vW_{j,i}^{(L)}\right] \E\left[\vphi_i^{(L)} \right] =\0,
    \sum_{j=1}^{C\ssup{\ell}} \ \sum_{\nu=1}^{H\ssup{\ell}\!D\ssup{\ell}} \E\sqb{W\ssup{\ell+1}_{i,j,\mu,\nu} \, \phi(A\ssup{\ell}_{j,\nu}(\vX))} = 0,
\end{equation*}
as $b_i\ssup{\ell+1}$ and $W\ssup{\ell+1}_{i,j,\mu,\nu}$ have zero mean, and $W\ssup{\ell+1}_{i,j,\nu,\mu}$ are independent of the activations at the previous layer, $\phi(A\ssup{\ell}_{j,\nu}(\vX))$.

Now we show that it is possible to efficiently compute the covariance function.
This is surprising because for many networks, we need to compute the covariance of activations between all pairs of locations in the feature map (i.e. $\Cov \sqb{A_{i,\mu}\ssup{\ell+1}(\vX), A_{i,\mu'}\ssup{\ell+1}(\vX')}$ for $\mu,\mu'\in\{1,\dotsc,H\ssup{\ell+1}D\ssup{\ell+1}\}$) and this object is extremely high-dimensional, $N^2 (H\ssup{\ell+1} D\ssup{\ell+1})^2$.
However, it turns out that we only need to consider the ``diagonal'' covariance, (i.e. we only need $\Cov \sqb{A_{i,\mu}\ssup{\ell+1}(\vX), A_{i,\mu}\ssup{\ell+1}(\vX')}$ for $\mu\in\{1,\dotsc,H\ssup{\ell+1}D\ssup{\ell+1}\}$), which is a more manageable quantity of size $N^2 (H\ssup{\ell+1} D\ssup{\ell+1})$.

This is true at the output layer $(L+1)$: in order to achieve an output
suitable for classification or regression, we use only a single output location
$H\ssup{L+1} = D\ssup{L+1} =1$, with a number of ``channels'' equal to the number of of outputs/classes, so it is only possible to compute the
covariance at that single location.
We now show that,
if we only need the covariance at corresponding locations in the outputs, we only need the covariance at corresponding locations in the inputs, and this requirement propagates backwards through the network.

Formally, as the activations are composed of a sum of terms, their covariance is the sum of the covariances of all those underlying terms,
\begin{equation}
  \begin{aligned}
&\Cov \sqb{A_{i, \mu}\ssup{\ell+1}(\vX), A_{i, \mu}\ssup{\ell+1}(\vX')} = 
\Var\sqb{b_i\ssup{\ell}} + \\
&\hspace{3em}
  \sum_{j=1}^{C\ssup{\ell}} \sum_{j'=1}^{C\ssup{\ell}} \ \sum_{\nu=1}^{H\ssup{\ell}\!D\ssup{\ell}} \ \sum_{\nu'=1}^{H\ssup{\ell}\!D\ssup{\ell}} \Cov\sqb{W\ssup{\ell+1}_{i,j,\mu,\nu} \phi(A\ssup{\ell}_{j,\nu}(\vX)), W\ssup{\ell+1}_{i,j',\mu,\nu'} \phi(A\ssup{\ell}_{j',\nu'}(\vX'))}.
  \end{aligned}
\end{equation}
As the terms in the covariance have mean zero, and as the weights and activations from the previous layer are independent,
\begin{equation}
  \begin{aligned}
&\Cov \sqb{A_{i, \mu}\ssup{\ell+1}(\vX), A_{i, \mu}\ssup{\ell+1}(\vX')} = 
  \sigma_\text{b}^2 + \\
&\hspace{3em}  \sum_{j=1}^{C\ssup{\ell}} \sum_{j'=1}^{C\ssup{\ell}} \ \sum_{\nu=1}^{H\ssup{\ell}\!D\ssup{\ell}} \ \sum_{\nu'=1}^{H\ssup{\ell}\!D\ssup{\ell}} 
    \E\sqb{W\ssup{\ell+1}_{i,j,\mu,\nu} W\ssup{\ell+1}_{i,j',\mu,\nu'}}
    \E\sqb{\phi(A\ssup{\ell}_{j,\nu}(\vX)) \phi(A\ssup{\ell}_{j',\nu'}(\vX'))}.
  \end{aligned}
\end{equation}
The weights are independent for different channels: $\vW\ssup{\ell+1}_{i,j}$
and $\vW\ssup{\ell+1}_{i,j'}$ are \ac{iid} for $j\neq j'$, so $\E\sqb{W\ssup{\ell+1}_{i,j,\mu,\nu} W\ssup{\ell+1}_{i,j',\mu,\nu'}} = 0$ for $j\neq j'$.
Further, each row $\mu$ of the weight matrices $\vW\ssup{\ell+1}_{i,j}$ only contains
independent variables or zeros (Fig.~\ref{fig:conv-as-linear}), so
$\E\sqb{W\ssup{\ell+1}_{i,j,\mu,\nu} W\ssup{\ell+1}_{i,j',\mu,\nu'}} = 0$ for $\nu \neq
\nu'$. Thus, we can eliminate the sums over $j'$ and $\nu'$:
\begin{equation}
  \Cov \sqb{A_{i, \mu}\ssup{\ell+1}(\vX), A_{i, \mu}\ssup{\ell+1}(\vX')} = 
  \sigma_\text{b}^2 +
  \sum_{j=1}^{C\ssup{\ell}} \ \sum_{\nu=1}^{H\ssup{\ell}\!D\ssup{\ell}}
    \E\sqb{W\ssup{\ell+1}_{i,j,\mu,\nu} W\ssup{\ell+1}_{i,j,\mu,\nu}}
    \E\sqb{\phi(A\ssup{\ell}_{j,\nu}(\vX)) \phi(A\ssup{\ell}_{j,\nu}(\vX'))}.
\end{equation}
% However, note that this is a specific property of the shape of the
% convolutional filters with the same output location: in general
% $W\ssup{\ell}_{j,i,g,h}$ and $W\ssup{\ell}_{j,i,g',h'}$ are not \ac{iid}, as the
% underlying convolutional filters imply that some of the weights, denoted
% $\vW\ssup{\ell}_{ji}$, are shared.

The $\mu$th row of $\vW\ssup{\ell+1}_{i,j}$ is zero for indices $\nu$ that do not
belong to its convolutional patch, so we can restrict the sum over $\nu$ to that
region.
We also define $v_g\ssup{1}(\vX, \vX')$, to emphasise that the covariances are independent of the output channel, $j$.
The variance of the first layer is
\begin{align}
  \label{eq:kernel-base}
  K_\mu\ssup{1}(\vX, \vX') =
   \Cov \sqb{A_{i, \mu}\ssup{1}(\vX), A_{i, \mu}\ssup{1}(\vX')} &= 
    \sigma_\text{b}^2 +
    %\Var\sqb{b_j\ssup{\ell+1}} +
    \frac{\sigma_\text{w}^2}{C\ssup{0}} \sum_{i=1}^{C\ssup{0}} \;\sum_{\nu\in \mu\text{th patch}}
    X_{i,\nu}  X'_{i,\nu}.
  \intertext{And we do the same for the other layers,}
  \label{eq:kernel-recursive}
  K_\mu\ssup{\ell+1}(\vX, \vX') =
  \Cov \sqb{A_{i, \mu}\ssup{\ell+1}(\vX), A_{i, \mu}\ssup{\ell+1}(\vX')} &=
  \sigma_\text{b}^2 +
  \sigma_\text{w}^2 \sum_{\nu\in \mu\text{th patch}} V_\nu\ssup{\ell}(\vX, \vX'),
    %\E\sqb{\phi(A\ssup{\ell}_{i,h}(\vX)) \phi(A\ssup{\ell}_{i,h}(\vX'))}.
\end{align}
where
\begin{equation}
  \label{eq:phi-cov}
  V_\nu\ssup{\ell}(\vX, \vX') = \E\sqb{\phi(A\ssup{\ell}_{j,\nu}(\vX)) \phi(A\ssup{\ell}_{j,\nu}(\vX'))}
\end{equation}
is the covariance of the activations, which is again independent of the channel.

\subsection{Covariance of the activities}

The elementwise covariance in the right-hand side of Eq.~\eqref{eq:kernel-recursive} can be computed
in closed form for many choices of $\phi$ if the activations are Gaussian. For each element of the
activations, one needs to keep track of the 3 distinct entries of the bivariate
covariance matrix between the inputs, 
$K_\mu\ssup{\ell+1}(\vX, \vX)$, $K_\mu\ssup{\ell+1}(\vX, \vX')$ and $K_\mu\ssup{\ell+1}(\vX', \vX')$.

For example, for the ReLU nonlinearity ($\phi(x) = \max(0, x)$), one can adapt \citet{cho2009kernel} in
the same way as \citet[section~3]{gpbehaviour} to obtain
\begin{equation}
  %s_g\ssup{\ell}(\vX, \vX') = \frac{\sqrt{\diagcov_{\vX\vX}\diagcov_{\vX'\vX'}}}{\pi} \left(\sin \theta + (\pi - \theta) \cos \theta\right)
  V_\nu\ssup{\ell}(\vX, \vX') = \frac{\sqrt{K_\nu\ssup{\ell}(\vX,\vX)K_\nu\ssup{\ell}(\vX',\vX')}}{\pi} \bra{\sin \theta_\nu\ssup{\ell} + (\pi - \theta_\nu\ssup{\ell}) \cos \theta_\nu\ssup{\ell}}
  \label{eq:nlin-relu}
\end{equation}
where $\theta_\nu\ssup{\ell} = \cos^{-1}\left( K_\nu\ssup{\ell}(\vX,\vX') /
  \sqrt{K_\nu\ssup{\ell}(\vX,\vX)K_\nu\ssup{\ell}(\vX',\vX')}\right)$.
Another
example is the error function (erf) nonlinearity, similar to the
hyperbolic tangent (tanh). The form of its relevant expectation
\citep{williams1997computing} is in Appendix~\ref{sec:erf}.

\begin{algorithm}[tb]
\begin{algorithmic}[1]
%\STATE \emph{Input}: two images, $\vX,\vX' \in \mathbb{R}^{C^{(0)} \times (H^{(0)}W^{(0)})}$.
%\STATE Compute $\bdiagcov_{\vX\vX}^{(1)}$, $\bdiagcov_{\vX\vX'}^{(1)}$, and
%$\bdiagcov_{\vX'\vX'}^{(1)}$; using \fref{eq:kernel-base}.
%\FOR{$\ell=1,2,\dots,L$}
%\STATE Compute $\E[\phi\phi]^{(\ell)}$, $\E[\phi\phi']^{(\ell)}$ and
%$\E[\phi'\phi']^{(\ell)}$ using equation (\ref{eq:nlin-relu}),
%(\ref{eq:nlin-erf}), or some other nonlinearity.
%\STATE Compute $\bdiagcov_{\vX\vX}^{(\ell+1)}$, $\bdiagcov_{\vX\vX'}^{(\ell+1)}$, and
%$\bdiagcov_{\vX'\vX'}^{(\ell+1)}$; using \fref{eq:kernel-recursive}.
%\ENDFOR
%\STATE Output $\diagcov_{\vX\vX'}^{(L+1)}$, which is a nonnegative scalar.
\STATE \emph{Input}: two images, $\vX,\vX' \in \mathbb{R}^{C^{(0)} \times (H^{(0)}W^{(0)})}$.
\STATE Compute $K_\mu\ssup{1}(\vX, \vX)$, $K_\mu\ssup{1}(\vX, \vX')$, and
$K_\mu\ssup{1}(\vX', \vX')$ \\\hspace{2em}for $\mu \in \{1,\dotsc,H\ssup{1}D\ssup{1}\}$; using Eq.~\eqref{eq:kernel-base}.
\FOR{$\ell=1,2,\dots,L$}
\STATE Compute $V_\mu\ssup{\ell}(\vX, \vX')$, $V_\mu\ssup{\ell}(\vX, \vX')$ and
$V_\mu\ssup{\ell}(\vX,\vX')$ \\\hspace{2em}for $\mu \in \{1,\dotsc,H\ssup{\ell}D\ssup{\ell}\}$; using Eq.~\eqref{eq:nlin-relu}, or some other nonlinearity.
\STATE Compute $K_\mu\ssup{\ell+1}(\vX,\vX)$, $K_\mu\ssup{\ell+1}(\vX,\vX')$, and
$K_\mu\ssup{\ell+1}(\vX',\vX')$ \\\hspace{2em}for $\mu \in \{1,\dotsc,H\ssup{\ell+1}D\ssup{\ell+1}\}$; using Eq.~\eqref{eq:kernel-recursive}.
\ENDFOR
\STATE Output the scalar $K_1\ssup{L+1}(\vX,\vX')$.
\end{algorithmic}
\caption{The ConvNet kernel $k(\vX, \vX')$}
\label{alg:kernel}
\end{algorithm}

\subsection{Efficiency of the ConvNet kernel\label{sec:efficiency}}

We now have all the pieces for computing the kernel, as written in
Algorithm~\ref{alg:kernel}.

Putting together Eq.~\eqref{eq:kernel-recursive} and 
Eq.~\eqref{eq:nlin-relu} gives us the surprising result that the diagonal
covariances of the activations at layer $\ell+1$ only depend on the 
diagonal covariances of the activations at layer $\ell$. This is very important,
because it makes the computational cost of the kernel be within a constant
factor of the cost of a forward pass for the equivalent \ac{CNN} with 1 filter
per layer.

Thus, the algorithm is more efficient
that one would naively think. A priori, one needs to compute the covariance
between all the elements of $\va_{j}\ssup{\ell}(\vX)$ and
$\va_{j}\ssup{\ell}(\vX')$ combined, yielding a $2H\ssup{\ell}D\ssup{\ell} \times
2H\ssup{\ell}D\ssup{\ell}$ covariance matrix for every pair of points. Instead, we
only need to keep track of a $H\ssup{\ell}D\ssup{\ell}$-dimensional vector per layer
and pair of points. %, and an amortised $H\ssup{\ell}D\ssup{\ell}$-dimensional vector per
%\emph{point} and layer.

Furthermore, the particular form for the kernel (Eq.~\ref{eq:network-base} and Eq.~\ref{eq:network-recursive}) implies that the required variances and covariances at all required locations can be computed efficiently as a convolution.

%Computing the kernel for a pair of points
%takes three \ac{CNN} forward passes, using equation (\ref{eq:nlin-relu}) or
%(\ref{eq:nlin-erf}) as the nonlinearity. Computing the kernel matrix for many
%points amortises the variances $\bdiagcov_{\vX\vX}^{(\ell)}$ and
%$\bdiagcov_{\vX'\vX'}^{(\ell)}$, and needs roughly 1 \ac{CNN} forward pass per
%pair of points.

\subsection{Kernel for a residual \ac{CNN}}

The induction step in the argument for \ac{GP} behaviour from
Sec.~\ref{sec:argument-gp} depends only on the previous activations being \ac{iid}
Gaussian. Since all the activations are \ac{iid} Gaussian,
we can add skip connections between the activations of different layers while preserving \ac{GP} behaviour,
e.g. $\vA^{(\ell+1)}$ and $\vA^{(\ell-s)}$ where $s$ is the number of layers that the skip
connection spans. If we change the \ac{NN} recursion
(Eq.~\ref{eq:network-recursive}) to
\begin{equation}
  \va_i^{(\ell+1)}(\vX) := \va_i^{(\ell-s)}(\vX) + \vb_i^{(\ell+1)} + \sum_{j=1}^{C^{(\ell)}}  \vW_{i,j}^{(\ell)} \phi\left( \va_j^{(\ell)}(\vX) \right),
\label{eq:resnet-recursive}
\end{equation}
then the kernel recursion (Eq.~\ref{eq:kernel-recursive}) becomes
\begin{equation}
  \label{eq:rekernel-recursive}
  K_\mu\ssup{\ell+1}(\vX, \vX') = K_\mu\ssup{\ell-s}(\vX, \vX')
    + \sigma_\text{b}^2 + \sigma_\text{w}^2
    \sum_{\nu \in \text{$\mu$th patch}} V_\nu\ssup{\ell}(\vX, \vX').
\end{equation}
This way of adding skip connections is equivalent to the ``pre-activation''
shortcuts described by \citet{he2016identity}. Remarkably, the
natural way of adding residual connections to \acp{NN} is the one that performed
best in their empirical evaluations.

\section{Experiments}

We evaluate our kernel on the MNIST handwritten digit classification task.
Classification likelihoods are not conjugate for \acp{GP}, so we must make an approximation, and we 
follow \citet{lee2017deep}, in re-framing classification as multi-output regression.

The training set is split into $N=50000$ training and $10000$ validation examples.
The regression targets $\vY \in \{-1, 1\}^{N \times 10}$ are a one-hot encoding
of the example's class: $y_{n,c} = 1$ if the $n$th example belongs to class $c$,
and $-1$ otherwise.

Training is exact conjugate likelihood \ac{GP} regression with noiseless
targets $\vY$
\citep{rasmussen2006gaussian}. First we compute the
$N\times N$ kernel matrix $\vK_{xx}$, which contains the
kernel between every pair of examples. Then we compute $\vK_{xx}^{-1}\vY$ using
a linear system solver.

The test set has $N_T=10000$ examples. We compute the $N_T \times N$ matrix
$\vK_{x^*x}$, the kernel between each test example and all the training examples.
The predictions are given by the row-wise maximum of
$\vK_{x^*x}\vK_{xx}^{-1}\vY$.

For the ``ConvNet \ac{GP}'' and ``Residual \ac{CNN} \ac{GP}'', (Table~\ref{mnist-results})
we optimise the kernel hyperparameters by random search. We draw $M$ random
hyperparameter samples, compute the resulting kernel's performance in the
validation set, and pick the highest performing run. The kernel hyperparameters
are: $\sigma_\text{b}^2$, $\sigma_\text{w}^2$; the number of layers; the convolution stride,
filter sizes and edge behaviour; the nonlinearity (we consider the error
function and ReLU); and the frequency of residual skip connections (for Residual
CNN GPs).
We do not retrain the model on the validation set after choosing hyperparameters.

The ``ResNet \ac{GP}'' (Table~\ref{mnist-results}) is the kernel equivalent to a 32-layer version of the
basic residual architecture by \citet{he2016deep}. The differences are: an initial $3
\times 3$ convolutional layer and a final dense layer instead
of average pooling.
We chose to remove the pooling because computing its output variance requires
the off-diagonal elements of the filter covariance, in which case we could not exploit the
efficiency gains described in Sec.~\ref{sec:efficiency}. %See \fref{sec:future-work}
%for a more detailed explanation.
% the output covariance in that case requires the covariance within a filter,
% removing the effiicenc

We found that the despite it not being optimised, the 32-layer ResNet \ac{GP} outperformed all other comparable architectures (Table~\ref{mnist-results}), including the NNGP in \citet{lee2017deep}, which is state-of-the-art for non-convolutional networks, and convolutional \acp{GP} \citep{markvdw2017convolutional,kumar2018deep}.
That said, our results have not reached state-of-the-art for methods that incorporate a parametric neural network, such as a standard ResNet \citep{chen2018neural} and a Gaussian process with a deep neural network kernel \citep{bradshaw2017adversarial}.

\begin{table}%[ht]
\begin{center}
\begin{tabular}{lcccc}
{\bf Method}  &{\bf \#samples} & {\bf Validation error} & {\bf Test error}  \\
  \hline
NNGP \citep{lee2017deep} & $\approx 250$  & -- & 1.21\% \\
Convolutional \ac{GP} \citep{markvdw2017convolutional} & SGD & -- & 1.17\% \\
Deep Conv. \ac{GP} \citep{kumar2018deep} & SGD & -- & 1.34\% \\
ConvNet \ac{GP} & 27 & 0.71\% & 1.03\% \\
Residual CNN \ac{GP} & 27 & 0.71\% & 0.93\% \\
ResNet \ac{GP} & -- & {0.68\%} & {\bf 0.84\%} \\
  \hline
\ac{GP} + parametric deep kernel \citep{bradshaw2017adversarial} & SGD & -- & 0.60\% \\
ResNet \citep{chen2018neural} & -- & -- & {\bf 0.41\%} \\
\end{tabular}
\end{center}
\caption{MNIST classification results. \#samples gives the number of kernels that were randomly
  sampled for the hyperparameter search. ``ConvNet GP'' and ``Residual CNN GP'' are
  random CNN architectures with a fixed filter size, whereas ``ResNet GP'' is a
  slight modification of
  the architecture by \citet{he2016identity}.
  Entries labelled ``SGD'' used stochastic gradient
  descent for tuning hyperparameters, by maximising the likelihood of the training set.
  The last two methods use parametric neural networks. The hyperparameters of
  the ResNet GP were not optimised \citep[they were fixed based on the
  architecture from][]{he2016identity}. See Table~\ref{table:hyperparameters} (appendix)
  for optimised hyperparameter values.%, and this GP performs
  %better than all other nonparametric approaches, but is still not as good as
  %methods with many parameters (below the line).  Note that the most directly comparable approach (and the state-of-the-art for non-convolutional architectures) is the NNGP \citep{lee2017deep}.
 \label{mnist-results}}

  %The Fully-Connected GP
%  used 50k and not 55k training examples.
%}
\end{table}

To check whether the GP limit is applicable to relatively small networks used
practically (with of the order of $100$ channels in the first layers), we
randomly sampled $10,000$ 32-layer ResNets, with 3, 10, 30 and 100 channels in
the first layers. Following the usual practice for ResNets we increase the
number the number of hidden units when we downsample the feature maps. Then, we
compare the sampled and limiting theoretical distributions of
$\vA\ssup{32}(\vX)$ for a given input $\vX$.

\begin{figure}[t]
  \includegraphics{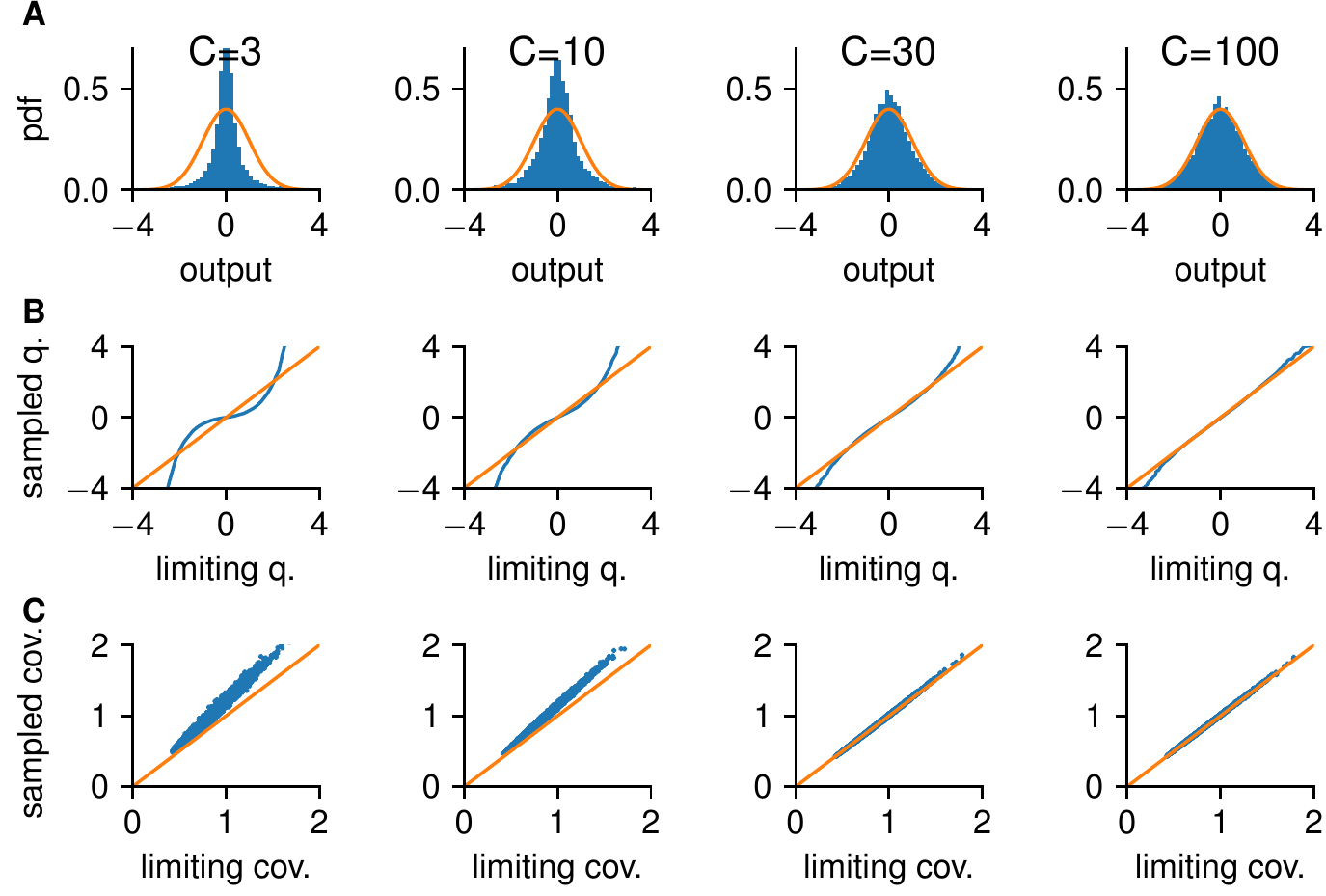}
  \caption{
Distribution of the output  
$\vA\ssup{32}(\vX)$:
limiting density and samples of finite 32-layer ResNets
\citep{he2016identity} with $C=3,10,30,100$ channels.
    % Comparison of the infinite limit, and outputs from finite 32-layer ResNets with 3, 10, 30, and 100 channels in their first layers.
    \textbf{A)} Empirical histogram and limiting density function for one input image.
    \textbf{B)} A more sensitive test of Gaussianity is a quantile-quantile plot,
    which plots in $x$ the value of a quantile in the limiting density and in
    $y$ the corresponding quantile in the empirical one, for one input image.
    \textbf{C)} The empirical moments (variances and covariances) over 100 training images show a good match for all numbers of channels.
    \label{fig:random}
  }
\end{figure}

The probability density plots show a good match around 100 channels (Fig.~\ref{fig:random}A), which matches a more sensitive graphical procedure based on quantile-quantile plots (Fig.~\ref{fig:random}B).
Notably, even for only 30 channels, the empirical moments (computed over many input
images) match closely the limiting ones (Fig.~\ref{fig:random}C).
For comparison, typical ResNets use from 64 \citep{he2016deep} to 192 \citep{zagoruyko2016wide} channels in their first layers.
We believe that this is because the moment propagation equations only require the Gaussianity assumption for propagation through the ReLU, and presumably this is robust to non-Gaussian input activations.

\paragraph{Computational efficiency.} 
Asymptotically, computing the kernel matrix takes $O(N^2 LD)$ time, where $L$ is the number of layers in the network and $D$ is the dimensionality of the input, and inverting the kernel matrix takes $O(N^3)$.
As such, we expect that for very large datasets, inverting the kernel matrix will dominate the computation time.
However, on MNIST, $N^3$ is only around a factor of $10$ larger than $N^2 LD$.
In practice, we found that it was more expensive to \emph{compute} the kernel matrix
than to invert it. For the ResNet kernel, the most expensive, computing $\vK_{xx}$, and $\vK_{xx*}$ for validation and test
took $3$h~$40$min on two Tesla~P100 GPUs. 
In contrast, inverting $\vK_{xx}$ and computing validation and test performance
took $43.25\pm 8.8$~seconds on a single Tesla~P100 GPU.
%Inverting the kernel matrix takes $O(N^3)$ time, which is higher. 
%the time for training our models was
%completely dominated by the time it took to compute the kernel matrix, even
%though this computation was run on GPU and the matrix inversion was run on CPU.

\section{Related work}

Van der Wilk et al. \citep{markvdw2017convolutional} also adapted \acp{GP} to
image classification.
They defined a prior on functions $f$ that takes an image and outputs a scalar. 
First, draw a function $g \sim \mathcal{GP}(0, k_p(\vX, \vX'))$. Then, $f$ is
the sum of the output of $g$ applied to each of the convolutional patches. %The
%result is that $f$ also follows a \ac{GP}, with kernel $k(\vX, \vX') \eqdef
%\sum_{i \in \text{patches}} \sum_{j \in \text{patches}} k_p(\vX_i, \vX'_j)$.
Their approach is also inspired by convolutional \acp{NN}, but their kernel $k_p$ is applied
to all pairs of patches of $\vX$ and $\vX'$. This makes their convolutional
kernel expensive to evaluate, requiring inter-domain inducing point
approximations to remain tractable.
The kernels in this
work, 
directly motivated by the infinite-filter limit of a \ac{CNN}, only
apply something like $k_p$ to the \emph{corresponding} pairs of patches within
$\vX$ and $\vX'$ (Eq.~\ref{eq:kernel-base}).
As such, the \ac{CNN} kernels are cheaper to compute and exhibit superior performance (Table~\ref{mnist-results}), despite the use of an approximate likelihood function.

\citet{kumar2018deep} define a prior over functions by stacking several \acp{GP}
with van der Wilk's convolutional kernel, forming a ``Deep \ac{GP}''
\citep{damianou13a}. In contrast, the kernel in this paper confines all hierarchy to the
definition of the kernel, and the resulting \acp{GP} is shallow.

\citet{wilson2016deep} introduced and \citet{bradshaw2017adversarial} improved deep kernel learning.
The inputs to a classic \ac{GP} kernel $k$ (e.g. RBF) are preprocessed by
applying a feature extractor $g$ (a deep \ac{NN}) prior to computing the kernel:
$k_\text{deep}(\vX, \vX') \eqdef k(g(\vX; \theta), g(\vX', \theta))$.
The \ac{NN} parameters are optimised by gradient ascent using the likelihood as the objective, as in standard \ac{GP} kernel learning \citep[Chapter~5]{rasmussen2006gaussian}. 
Since deep kernel learning incorporates a state-of-the-art \ac{NN} with over
$10^6$ parameters, we expect it to perform similarly to a \ac{NN} applied
directly to the task of image classification. At present both \acp{CNN} and deep
kernel learning
display superior performance to the \ac{GP} kernels in this work.
%As such, at present both a \ac{NN} applied directly to the task of image classification, and a deep-kernel have superior performance 
However, the kernels defined here have far fewer parameters (around $10$,
compared to their $10^6$).
%
% We expect the performance gap to close as we
% incorporate additional features such as pooling (which requires computing
% off-diagonal elements of the covariance in \fref{sec:gp-covariance})
% or improved
% approximations to the true likelihood function.
%
%However, the very large number of parameters in the \ac{NN} (over $10^6$ in comparison to our two comparable parameters $\sigma_b^2$ and $\sigma_w^2$),  reintroduces the risk of overfitting, and as such sacrifices the benefits of exact Bayesian inference in \acp{GP}.
%That said, state-of-the-art neural networks applied directly to MNIST classification at present give superior performance to our \ac{GP} methods. 
%As such, we expect (and do indeed find) that \acp{GP} that use the outputs of such a \ac{NN} (improved by \citet{bradshaw2017adversarial}) also exhibit superior performance.
%However, we expect this gap to close as we 
%That said, the use of a full neural network allows for superior 
%the approach was improved by \citet{bradshaw2017adversarial} and performs better than ours. 
%Their approach was improved by \citet{bradshaw2017adversarial} and performs better than ours. 
%However, their kernel has more than $10^6$ parameters (ignoring architectural choices), whereas our kernel has 2 comparable parameters, $\sigma_b^2$ and $\sigma_w^2$. 
%Also unlike us, they use a proper classification likelihood function.
% The work by Mairal on convolutional kernels.

\citet{borovykh} also suggests that a \ac{CNN} exhibits \ac{GP} behaviour. 
However, they take the infinite limit with respect to the \emph{filter size}, not the number of filters. 
Thus, their infinite network is inapplicable to real data which is always of finite dimension.

Finally, there is a series of papers analysing the mean-field behaviour of deep \acp{NN} and \acp{CNN} which aims to find good random initializations, i.e.\ those that do not exhibit vanishing or exploding gradients or activations \citep{schoenholz2016deep,yang2017mean}.
Apart from their very different focus, the key difference to our work is that they compute the variance for a single training-example, whereas to obtain the \acp{GP} kernel, we additionally need to compute the output covariances for different training/test examples \citep{xiao2018dynamical}.

\section{Conclusions and future work}

We have shown that deep Bayesian \acp{CNN} with infinitely many filters are
equivalent to a \ac{GP} with a recursive kernel. We also derived the kernel for the \ac{GP} equivalent to
a \ac{CNN}, and showed that, in handwritten digit classification, it outperforms
all previous \ac{GP} approaches that do not incorporate a parametric \ac{NN} into the
kernel.
Given that most state-of-the-art neural networks
incorporate structure (convolutional or otherwise) into their architecture,
the equivalence between \acp{CNN} and \acp{GP} is potentially of
considerable practical relevance.
In particular, we hope to apply \ac{GP} \acp{CNN} in domains as widespread as adversarial examples, lifelong learning and k-shot learning, and we hope to improve them by developing efficient multi-layered inducing point approximation schemes.

%\subsubsection*{Acknowledgments}
%We wish to thank Will Tebbutt, Richard Turner, and others, for helpful comments and feedback on this work.
%This work was funded by ... and the Howard Hughes Medical Institute.

\bibliography{already-read}
\bibliographystyle{iclr2019_conference}

\section{Appendix}

\subsection{Technical notes on limits\label{sec:notes-limits}}

The key technical issues in the proof (and the key differences between \citealt{lee2017deep} and \citealt{matthews2018gaussian}) arise from exactly how and where we take limits.
In particular, consider the activations as being functions of the activities at the previous layer,
\begin{align}
  \vA\ssup{4} &= \vA\ssup{4}(\vA\ssup{3}(\vA\ssup{2}(\vA\ssup{1}(\vX))))
\end{align}
Now, there are two approaches to taking limits.
First, both our argument in the main text, and the argument in \citet{lee2017deep} is valid if we are able to take limits ``inside'' the network,
\begin{align}
  \vA\ssup{4}_\text{L} &= \lim_{C\ssup{3}\rightarrow \infty} \vA\ssup{4} \bra{\lim_{C\ssup{2}\rightarrow \infty}\vA\ssup{3}\bra{\lim_{C\ssup{1}\rightarrow \infty} \vA\ssup{2}\bra{\vA\ssup{1}(\vX)}}}.
\end{align}
However, \citet{gpbehaviour,matthews2018gaussian} argue that is preferable to take limits ``outside'' the network.
In particular, \citet{matthews2018gaussian} take the limit with all layers simultaneously,
\begin{align}
  \vA\ssup{4}_\text{M} &= \lim_{n\rightarrow \infty} \vA\ssup{4} \bra{\vA\ssup{3}\bra{ \vA\ssup{2}\bra{\vA\ssup{1}(\vX)}}},
\end{align}
where $C\ssup{\ell} = C\ssup{\ell}(n)$ goes to infinity as $n \rightarrow \infty$. That said, similar technical issues arise if we take limits in sequence, but outside the network.

\subsection{Extending the derivations of \citet{matthews2018gaussian} to the convolutional case\label{sec:actual-proof}}

In the main text, we follow \citet{lee2017deep} in sequentially taking the limit of each layer to infinity (i.e. $C\ssup{1}\rightarrow \infty$, then $C\ssup{2}\rightarrow \infty$ etc.).
This dramatically simplified the argument, because taking the number of units in the previous layer to infinity means that the inputs from that layer are exactly Gaussian distributed.
However, \citet{matthews2018gaussian} argue that the more practically relevant limit is where we take all layers to infinity simultaneously.
This raises considerable additional difficulties, because we must reason about convergence in the case where the previous layer is finite.
Note that this section is not intended to stand independently: it is intended to be read alongside \citet{matthews2018gaussian}, and we use several of their results without proof.

Mirroring Definition 3 in \citet{matthews2018gaussian}, we begin by choosing a set of ``width'' functions, $C\ssup{\ell}(n)$, for $\ell \in \{1,\dotsc,L\}$ which all approach infinity as $n \rightarrow \infty$.
In \citet{matthews2018gaussian}, these functions described the number of hidden units in each layer, whereas here they describe the number of channels.
Our goal is then to extend the proofs in \citet{matthews2018gaussian} (in particular, of theorem 4), to show that the output of our convolutional networks converge in distribution to a Gaussian process as $n \rightarrow \infty$, with mean zero and covariance given by the recursion in Eqs.~(\ref{eq:kernel-base} -- \ref{eq:phi-cov}).

The proof in \citet{matthews2018gaussian} has three main steps.
First, they use the Cramér-Wold device, to reduce the full problem to that of proving convergence of scalar random variables to a Gaussian with specified variance.
Second, if the previous layers have finite numbers of channels, then the channels $\va_j\ssup{\ell}(\vX)$ and $\va_j\ssup{\ell}(\vX')$ are uncorrelated but no longer independent, so we cannot apply the CLT directly, as we did in the main text.
Instead, they write the activations as a sum of exchangeable random variables, and derive an adapted CLT for exchangeable (rather than independent) random variables \citep{blum1958central}.
Third, they show that moment conditions required by their exchangeable CLT are satisfied.

%To extend their proofs we need to consider these three steps.
%First, we apply the Cramér-Wold device in the same fashion.
%Second, the key step is to define their projections and summands that allow us to apply the exchangeable CLT.
%Third, we consider only bounded activation functions, in which case the moment conditions simplify dramatically.
To extend their proofs to the convolutional case, we begin by defining our networks in a form that is easier to manipulate and as close as possible to Eq. (21-23) in \citet{matthews2018gaussian},
\begin{align}
  A_{i, \mu}\ssup{1} = f_{i, \mu}\ssup{1}(x) &= \frac{\sigma_\text{w}}{\sqrt{C\ssup{0}}} \sum_{j=1}^{C\ssup{0}} \sum_{\nu \in \text{$\mu$th patch}} \epsilon_{i,j,\mu,\nu}\ssup{1} x_{j, \nu} + b_i\ssup{1}, \quad i \in \mathbb{N}\\
  g_{i, \mu}\ssup{\ell}(x) &= \phi\bra{f_{i, \mu}\ssup{\ell}(x)}\\
  \label{eq:def:f:rec}
  A_{i, \mu}\ssup{\ell+1} = f_{i, \mu}\ssup{\ell+1}(x) &= \frac{\sigma_\text{w}}{\sqrt{C\ssup{\ell}(n)}} \sum_{j=1}^{C\ssup{\ell}(n)} \sum_{\nu \in \text{$\mu$th patch}} \epsilon_{i,j,\mu,\nu} \ssup{\ell+1} g_{j, \nu}\ssup{\ell}(x) + b_i\ssup{\ell+1}, \quad i \in \mathbb{N}
  \intertext{where,}
  \epsilon_{i,j,\mu,\nu} &\sim \mathcal{N}(0, 1).
\end{align}

The first step is to use the Cramér-Wold device \citep[Lemma 6 in][]{matthews2018gaussian}, which indicates that convergence in distribution of a sequence of finite-dimensional vectors is equivalent to convergence on all possible linear projections to the corresponding real-valued random variable.
Mirroring Eq.~25 in \citet{matthews2018gaussian}, we consider convergence of random vectors, $f\ssup{\ell}_{i,\mu}(x)[n] - b_{i}\ssup{\ell}$, projected onto $\alpha\ssup{x, i, \mu}$,
\begin{align}
  \mathcal{T}\ssup{\ell}\bra{\mathcal{L}, \alpha}[n] = \sum_{(x, i, \mu) \in \mathcal{L}} \alpha\ssup{x, i, \mu} \sqb{f\ssup{\ell}_{i,\mu}(x)[n] - b_{i}\ssup{\ell}}.
\end{align}
where $\mathcal{L} \subset \mathcal{X} \times \mathbb{N} \times \{1,\dotsc,H\ssup{\ell}D\ssup{\ell}\}$ is a finite set of tuples of data points and channel indicies, $i$, and indicies of elements within channels/feature maps, $\mu$.
The suffix $[n]$ indicates width functions that are instantiated with input, $n$.

Now, we must prove that these projections converge in distribution a Gaussian.
We begin by defining summands, as in Eq. 26 in \citet{matthews2018gaussian},
\begin{align}
  \gamma_j\ssup{\ell}\bra{\mathcal{L}, \alpha}[n] := \sigma_\text{w} \sum_{(x, i, \mu) \in \mathcal{L}} \alpha\ssup{x, i, \mu} \sum_{\nu \in \text{$\mu$th patch}} \epsilon_{i,j,\mu,\nu}\ssup{\ell} g_{j, \nu}\ssup{\ell-1}(x)[n],
\end{align}
such that the projections can be written as a sum of the summands, exactly as in Eq. 27 in \citet{matthews2018gaussian},
\begin{align}
  \mathcal{T}\ssup{\ell}\bra{\mathcal{L}, \alpha}[n] = \frac{1}{\sqrt{C\ssup{\ell-1}(n)}} \sum_{j=1}^{C\ssup{\ell-1}(n)} \gamma_j\ssup{\ell}\bra{\mathcal{L}, \alpha}[n].
\end{align}
Now we can apply the exchangeable CLT to prove that $\mathcal{T}\ssup{\ell}\bra{\mathcal{L}, \alpha}[n]$ converges to the limiting Gaussian implied by the recursions in the main text.
To apply the exchangeable CLT, the first step is to mirror Lemma 8 in \citet{matthews2018gaussian}, in showing that for each fixed $n$ and $\ell \in \{2,\dotsc,L+1\}$, the summands, $\gamma_j\ssup{\ell}\bra{\mathcal{L}, \alpha}[n]$ are exchangeable with respect to the index $j$.
In particular, we apply de Finetti's theorem, which states that a sequence of random variables is exchangeable if and only if they are i.i.d. conditional on some set of random variables, so it is sufficient to exhibit such a set of random variables.
Mirroring Eq. 29 in \citet{matthews2018gaussian}, we apply the recursion,
\begin{equation}
  \begin{aligned}
  \gamma_j\ssup{\ell}\bra{\mathcal{L}, \alpha}[n] :=  &\sigma_\text{w} \sum_{(x, i, \mu) \in \mathcal{L}} \alpha\ssup{x, i, \mu} \sum_{\nu \in \text{$\mu$th patch}} \epsilon_{i,j,\mu,\nu}\ssup{\ell}\\
  &\hspace{1em}\phi\bra{\frac{\sigma_\text{w}}{\sqrt{C\ssup{\ell-2}(n)}} \sum_{k=1}^{C\ssup{\ell-2}(n)} \sum_{\xi \in \text{$\nu$th patch}} \epsilon_{j,k,\nu,\xi}\ssup{\ell-1} g_{k, \xi}\ssup{\ell-2}(x)[n] + b_j\ssup{\ell+1}}
  \end{aligned}
\end{equation}
As such, the summands are iid conditional on the finite set of random variables $\left\{ g_{k, \xi}\ssup{\ell-2}(x)[n]: k\in\{1,\dotsc,C\ssup{\ell-2}\},\xi\in\{1,\dotsc,H\ssup{\ell-2}D\ssup{\ell-2}\}, x\in\mathcal{L}_\mathcal{X}\right\}$, where $\mathcal{L}_\mathcal{X}$ is the set of input points in $\mathcal{L}$.

The exchangeable CLT in Lemma 10 in \citet{matthews2018gaussian} indicates that $\mathcal{T}\ssup{\ell}\bra{\mathcal{L}, \alpha}[n]$ converges in distribution to $\mathcal{N}\bra{0, \sigma_*^2}$ if the summands are exchangeable (which we showed above), and if three conditions hold,
\renewcommand{\labelenumi}{\alph{enumi})}
\begin{enumerate}
  \item $\E_n\sqb{\gamma_j\ssup{\ell} \gamma_{j'}\ssup{\ell}} = 0$
  \item $\lim_{n\rightarrow \infty} \E_n\sqb{\bra{\gamma_j\ssup{\ell}}^2 \bra{\gamma_{j'}\ssup{\ell}}^2} = \sigma_*^4$
  \item $\E_n\sqb{|\gamma_j\ssup{\ell}|^3} = o\bra{\sqrt{C\ssup{\ell}(n)}}$
\end{enumerate}
Condition a) follows immediately as the summands are uncorrelated and zero-mean.
Conditions b) and c) are more involved as convergence in distribution in the previous layers does not imply convergence in moments for our activation functions.

We begin by considering the extension of Lemma 20 in \citet{matthews2018gaussian}, which allow us to show conditions b) and c) above, even in the case of unbounded but linearly enveloped nonlinearities \citep[Definition 1 in][]{matthews2018gaussian}.
Lemma 20 states that the eighth moments of $f_{i, \mu}\ssup{t}(x)[n]$ are bounded by a finite constant independent of $n \in \mathbb{N}$.
We prove this by induction.
The base case is trivial, as $f_{j, \mu}\ssup{1}(x)[n]$ is Gaussian.
Following \citet{matthews2018gaussian}, assume the condition holds up to $\ell-1$, and show that the condition holds for layer $\ell$.
Using Eq.~\eqref{eq:def:f:rec}, we can bound the activations at layer $\ell$,
\begin{align}
  \E\sqb{\lvert f_{i, \mu}\ssup{\ell}(x)[n] \rvert^8} \leq 2^{8-1} \E\sqb{\lvert b_i\ssup{\ell}\rvert^8 + \left\lvert \frac{\sigma_\text{w}}{\sqrt{C\ssup{\ell-1}}}\sum_{j=1}^{C\ssup{\ell-1}(n)} \sum_{\nu \in \text{$\mu$th patch}} \epsilon_{i,j,\mu,\nu}\ssup{\ell} g_{j, \nu}\ssup{\ell-1}(x)[n] \right\rvert^8}
\end{align}
Following Eq.~48 in \citet{matthews2018gaussian}, which uses Lemma 19 in \citet{matthews2018gaussian}, we have,
\begin{multline}
  \label{eq:bound}
  \E\sqb{\left\lvert \frac{\sigma_\text{w}}{\sqrt{C\ssup{\ell-1}}}\sum_{j=1}^{C\ssup{\ell-1}(n)} \sum_{\nu \in \text{$\mu$th patch}} \epsilon_{i,j,\mu,\nu}\ssup{\ell} g_{j, \nu}\ssup{\ell-1}(x)[n] \right\rvert^8} \\
  = \frac{2^4 \Gamma(4 + 1/2)}{\Gamma(1/2)} \E\sqb{\left\lvert \frac{\sigma_\text{w}^2}{C\ssup{\ell-1}(n)} \lVert g_{j\in\{1,\dotsc,C\ssup{\ell-1}(n)\},\nu\in\text{$\mu$th patch}}\ssup{\ell-1}(x)[n] \rVert_2^2 \right\rvert^4}.
\end{multline}
where $g_{j\in\{1,\dotsc,C\ssup{\ell-1}(n)\},\nu\in\text{$\mu$th patch}}\ssup{\ell-1}(x)[n]$ is the set of post-nonlinearities corresponding to $j\in\{1,\dotsc,C\ssup{\ell-1}(n)\}$ and $\nu\in\text{$\mu$th patch}$.
Following \citet{matthews2018gaussian}, observe that,
\begin{align}
  \frac{1}{C\ssup{\ell-1}(n)} \lVert g_{j\in\{1,\dotsc,C\ssup{\ell-1}(n)\},\nu\in\text{$\mu$th patch}}\ssup{\ell-1}(x)[n] \rVert_2^2
  &= \frac{1}{C\ssup{\ell-1}(n)} \sum_{j=1}^{C\ssup{\ell-1}(n)} \sum_{\nu \in \text{$\mu$th patch}} \bra{g_{j, \nu}\ssup{\ell-1}(x)[n]}^2\\
  &\hspace{-3em}\leq \frac{1}{C\ssup{\ell-1}(n)} \sum_{j=1}^{C\ssup{\ell-1}(n)} \sum_{\nu \in \text{$\mu$th patch}} \bra{c + m \lvert f_{j, \nu}\ssup{\ell-1}(x)[n] \rvert}^2
\end{align}
by the linear envelope property, $\lvert\phi(u)\rvert \leq c + m \lvert u\rvert$.  Following \citet{matthews2018gaussian}, we substitute this bound back into Eq.~\eqref{eq:bound} and suppress a multiplicative constant independent of $x$ and $n$,
\begin{multline}
  \E\sqb{\left\lvert \frac{\sigma_\text{w}}{\sqrt{C\ssup{\ell-1}(n)}}\sum_{j=1}^{C\ssup{\ell-1}(n)} \sum_{\nu \in \text{$\mu$th patch}} \epsilon_{i,j,\mu,\nu}\ssup{\ell} g_{j, \nu}\ssup{\ell-1}(x)[n] \right\rvert^8} \\
  \leq \frac{1}{\bra{C\ssup{\ell-1}(n)}^4} \E\sqb{\left\lvert \sum_{j=1}^{C\ssup{\ell-1}(n)} \sum_{\nu \in \text{$\mu$th patch}} c^2 + 2cm \lvert f_{j, \mu}\ssup{\ell-1}(x)[n] \rvert + m^2 \lvert f_{j, \mu}\ssup{\ell-1}(x)[n]\rvert^2 \right\rvert^4}
\end{multline}
This can be multiplied out, yielding a weighted sum of expectations of the form,
\begin{align}
  \E\sqb{
    \lvert f_{k,\nu}\ssup{\ell-1}(x)[n]\rvert^{p_1}
    \lvert f_{l,\xi}\ssup{\ell-1}(x)[n]\rvert^{p_2}
    \lvert f_{r,\pi}\ssup{\ell-1}(x)[n]\rvert^{p_3}
    \lvert f_{q,\rho}\ssup{\ell-1}(x)[n]\rvert^{p_4}
  }
\end{align}
with $p_i \in \{0, 1, 2\}$ for $i=1,2,3,4$, and $k,l,r,q\in \{1,\dotsc,C\ssup{\ell-1}(n)\}$, and $\nu,\xi,\pi,\rho\in\text{$\mu$th patch}$ where the weights of these terms are independent of $n$.
Using Lemma 18 in \citet{matthews2018gaussian}, each of these terms is bounded if the eighth moments of $f_{k,\mu}\ssup{\ell-1}(x)[n]$ are bounded, which is our inductive hypothesis.
The number of terms in the expanded sum is upper bounded by $\bra{2 C\ssup{\ell-1}(n) \lvert \text{$\mu$th patch} \rvert}^4$, where $\lvert \text{$\mu$th patch} \rvert$ is the number of elements in a convolutional patch.
Thus, we can use the same constant for any $n$ due to the $1/\bra{C\ssup{\ell-1}(n)}^4$ scaling.
As in \citet{matthews2018gaussian}, noting that $f_{j,\mu}\ssup{\ell-1}(x)[n]$ are exchangeable over $j$ for any $x$ and $n$ concludes the proof.

Using this result, we can obtain a straightforward adaptation of Lemmas 15, 16 and 21 in \citet{matthews2018gaussian}.
Lemma 15 gives condition b), Lemma 16 gives condition c); Lemma 15 requires uniform integrability, which is established by Lemma 21.

\subsection{Calibration of Gaussian process uncertainty}
\begin{figure}[h]
  \centerline{\includegraphics[width=\textwidth]{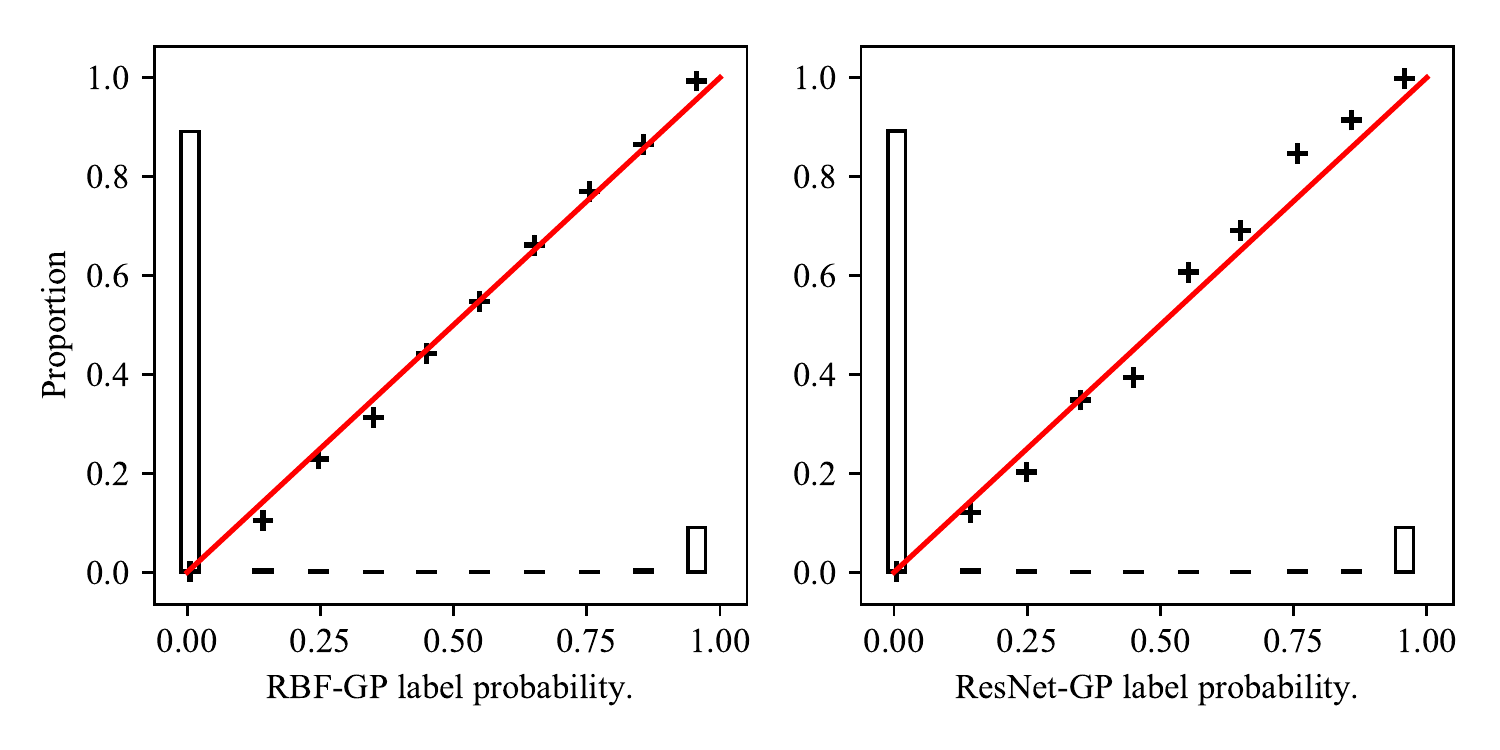}}
  \caption{
    Calibration plots for an RBF kernel (left) and the ResNet kernel (right). 
    The x-axis gives GP prediction for the label probability.
    The points give corresponding proportion of test points with that label, and the bars give the proportion of training examples in each bin.
    \label{fig:calibration}
  }
\end{figure}

It is important to check that the estimates of uncertainty produced by our Gaussian process are reasonable.
However, to make this assessment, we needed to use a proper likelihood, and not the squared-error loss in the main text.
We therefore used our kernel to perform the full, multi-class classification problem in GPflow \citep{matthews2017gpflow}, with a RobustMax likelihood \citep{robustmax}.
The more difficult non-conjugate inference problem forced us to use 1000 inducing points, randomly chosen from the training inputs.
Both our kernel and an RBF kernel have similar calibration curves, that closely track the diagonal, indicating accurate uncertainty estimation.
However, even in the inducing point setting, our convolutional kernel gave
considerably better performance than the RBF kernel (2.4\% error vs 3.4\%
error). See Fig.~\ref{fig:calibration}.

\subsection{Closed-form expectation for the error function nonlinearity\label{sec:erf}}

The error function (erf) is given by the integral
$\phi(x) = \frac{2}{\sqrt{\pi}}\int_0^xe^{-t^2}dt$, and is related to the
Gaussian CDF. It is very similar to the hyperbolic tangent (tanh), with the
difference that erf's tails approach 0 and 1 faster than the tanh's tails.

\citet{williams1997computing} gives us the relevant closed-form integral:
\begin{equation}
V_\nu\ssup{\ell}(\vX, \vX')
= \frac{2}{\pi} \sin^{-1}\left( \frac{2\,K_\nu\ssup{\ell}(\vX,\vX')}{\sqrt{(1 + 2\,K_\nu\ssup{\ell}(\vX,\vX)(1 + 2\,K_\nu\ssup{\ell}(\vX',\vX'))}}\right).
\label{eq:nlin-erf}
\end{equation}

\begin{table}[h]
\begin{center}
\begin{tabular}{lcccc}
{\bf Hyperparameters}  &{\bf ConvNet GP} & {\bf Residual CNN GP} & {\bf ResNet GP}  \\
  \hline
  $\sigma^2_\text{b}$ & 7.86 & 4.69 & 0.0 \\
  $\sigma^2_\text{w}$ & 2.79 & 7.27 & 1.0\\
  \#layers & 7 & 9 & 32 \\
  Stride & 1 & 1 & mostly 1, some 2 \\
  Filter sizes & 7 & 4 & 3 \\
  Padding & SAME & SAME & SAME \\
  Nonlinearity & ReLU & ReLU & ReLU \\
  Skip connections & -- & every 1 layer & every 2 layers \\
  \hline
  Test error & 1.03\% & 0.93\% & 0.84\% \\
\end{tabular}
\end{center}
\caption{Optimised hyperparameter values. The ResNet has $\sigma^2_\text{b}=0$
  because there are no biases in the architecture of \citet{he2016deep}. Because
  of this, and the fact that
  the nonlinearity is a ReLU, the value of $\sigma^2_\text{w}$ does not matter except
  for numerical stability: the $\sigma^2_\text{w}$ for every layer can be taken out of the
  nonlinearity and multiplied together, and kernel functions that are equal up
  to multiplication give the same classification results in this Gaussian
  likelihood setting.
  \label{table:hyperparameters}}
\end{table}

\end{document}

%% file: img/convolution.pdf_tex
%% Creator: Inkscape inkscape 0.92.2, www.inkscape.org
%% PDF/EPS/PS + LaTeX output extension by Johan Engelen, 2010
%% Accompanies image file 'convolution.pdf' (pdf, eps, ps)
%%
%% To include the image in your LaTeX document, write
%%   \input{<filename>.pdf_tex}
%%  instead of
%%   \includegraphics{<filename>.pdf}
%% To scale the image, write
%%   \def\svgwidth{<desired width>}
%%   \input{<filename>.pdf_tex}
%%  instead of
%%   \includegraphics[width=<desired width>]{<filename>.pdf}
%%
%% Images with a different path to the parent latex file can
%% be accessed with the `import' package (which may need to be
%% installed) using
%%   \usepackage{import}
%% in the preamble, and then including the image with
%%   \import{<path to file>}{<filename>.pdf_tex}
%% Alternatively, one can specify
%%   \graphicspath{{<path to file>/}}
%% 
%% For more information, please see info/svg-inkscape on CTAN:
%%   http://tug.ctan.org/tex-archive/info/svg-inkscape
%%
\begingroup%
  \makeatletter%
  \providecommand\color[2][]{%
    \errmessage{(Inkscape) Color is used for the text in Inkscape, but the package 'color.sty' is not loaded}%
    \renewcommand\color[2][]{}%
  }%
  \providecommand\transparent[1]{%
    \errmessage{(Inkscape) Transparency is used (non-zero) for the text in Inkscape, but the package 'transparent.sty' is not loaded}%
    \renewcommand\transparent[1]{}%
  }%
  \providecommand\rotatebox[2]{#2}%
  \ifx\svgwidth\undefined%
    \setlength{\unitlength}{233.5070945bp}%
    \ifx\svgscale\undefined%
      \relax%
    \else%
      \setlength{\unitlength}{\unitlength * \real{\svgscale}}%
    \fi%
  \else%
    \setlength{\unitlength}{\svgwidth}%
  \fi%
  \global\let\svgwidth\undefined%
  \global\let\svgscale\undefined%
  \makeatother%
  \begin{picture}(1,0.70071001)%
    \put(0.01422771,0.48371476){\color[rgb]{0,0,0}\makebox(0,0)[lb]{\smash{Filter $\vU\ssup{0}_{i,j}$:}}}%
    \put(0.00868537,0.6178629){\color[rgb]{0,0,0}\makebox(0,0)[lb]{\smash{Input image's $j$th channel:}}}%
    \put(0,0){\includegraphics[width=\unitlength,page=1]{convolution.pdf}}%
    \put(0.66275017,0.26523534){\color[rgb]{0,0,0}\makebox(0,0)[lb]{\smash{=}}}%
    \put(0,0){\includegraphics[width=\unitlength,page=2]{convolution.pdf}}%
    \put(0.73740812,0.57172831){\color[rgb]{0,0,0}\makebox(0,0)[lb]{\smash{Resulting}}}%
    \put(0.71775488,0.51661847){\color[rgb]{0,0,0}\makebox(0,0)[lb]{\smash{convolution}}}%
    \put(0,0){\includegraphics[width=\unitlength,page=3]{convolution.pdf}}%
    \put(0.28012076,0.06764961){\color[rgb]{0,0,0}\makebox(0,0)[lb]{\smash{$\vW\ssup{0}_{i,j}$}}}%
    \put(0.5829519,0.01211985){\color[rgb]{0,0,0}\makebox(0,0)[lb]{\smash{$\vx_j$}}}%
    \put(0.07409319,0.21129803){\color[rgb]{0,0,0}\rotatebox{90}{\makebox(0,0)[lb]{\smash{$\mu$th row}}}}%
    \put(0.90118064,0.16180928){\color[rgb]{0,0,0}\rotatebox{90}{\makebox(0,0)[lb]{\smash{$\mu$th conv. patch}}}}%
  \end{picture}%
\endgroup%